%% file: main.tex
\documentclass{article}

\usepackage{microtype}
\usepackage{float}
\usepackage{amsfonts} 
\usepackage{amsmath}
\usepackage{graphicx}
\usepackage{subfigure}
\usepackage{wrapfig}
\usepackage{caption}
\usepackage{multirow}
\usepackage{booktabs}
\usepackage{arydshln}

\usepackage{tikz}
\usetikzlibrary{bayesnet}
\tikzset{>=latex}
\usetikzlibrary{arrows,automata}

\usepackage{algorithm}
\usepackage{algorithmic}
\usepackage{booktabs} % for professional tables

\DeclareMathOperator*{\argmax}{arg\,max}

\usepackage{hyperref}

\usepackage[accepted]{icml2019}

\icmltitlerunning{Learning a Prior over Intent via Meta-Inverse Reinforcement Learning}

\begin{document}

\twocolumn[
\icmltitle{Learning a Prior over Intent\\via Meta-Inverse Reinforcement Learning}

% It is OKAY to include author information, even for blind
% submissions: the style file will automatically remove it for you
% unless you've provided the [accepted] option to the icml2019
% package.

% List of affiliations: The first argument should be a (short)
% identifier you will use later to specify author affiliations
% Academic affiliations should list Department, University, City, Region, Country
% Industry affiliations should list Company, City, Region, Country

% You can specify symbols, otherwise they are numbered in order.
% Ideally, you should not use this facility. Affiliations will be numbered
% in order of appearance and this is the preferred way.
%\icmlsetsymbol{equal}{*}

\begin{icmlauthorlist}
\icmlauthor{Kelvin Xu}{berkeley}
\icmlauthor{Ellis Ratner}{berkeley}
\icmlauthor{Anca Dragan}{berkeley}
\icmlauthor{Sergey Levine}{berkeley}
\icmlauthor{Chelsea Finn}{berkeley}
\end{icmlauthorlist}

\icmlaffiliation{berkeley}{Department of Electrical Engineering and Computer Science, University of California, Berkeley, USA}

\icmlcorrespondingauthor{Kelvin Xu}{kelvinxu@berkeley.edu}

% You may provide any keywords that you
% find helpful for describing your paper; these are used to populate
% the "keywords" metadata in the PDF but will not be shown in the document
\icmlkeywords{Machine Learning, ICML}

\vskip 0.3in
]

% this must go after the closing bracket ] following \twocolumn[ ...

% This command actually creates the footnote in the first column
% listing the affiliations and the copyright notice.
% The command takes one argument, which is text to display at the start of the footnote.
% The \icmlEqualContribution command is standard text for equal contribution.
% Remove it (just {}) if you do not need this facility.

\printAffiliationsAndNotice{}  % leave blank if no need to mention equal contribution
%\printAffiliationsAndNotice{\icmlEqualContribution} % otherwise use the standard text.

\input{defs.tex}
\input{abstract.tex}
\input{intro.tex}
\input{related_work.tex}
\input{prelim.tex}
\input{methods.tex}
\input{suncg_experiments.tex}
\input{discussion.tex}

%%%%%%%%%%%%%%%%%%%%%%%%%%%%%%%%%%%%%%%%%%%%%%%%%%%%%%%%%%%%%%%%%%%%%%%%%%%%%%%
%%%%%%%%%%%%%%%%%%%%%%%%%%%%%%%%%%%%%%%%%%%%%%%%%%%%%%%%%%%%%%%%%%%%%%%%%%%%%%%
\subsubsection*{Acknowledgments}
We thank Frederik Ebert,  Adam Gleave, Erin Grant, Sergio Guadarrama, Rowan McAllister, Charlotte Nguyen, Sid Reddy and Aravind Srinivas for comments on an earlier version of this paper. This work as supported by the Open Philanthropy Foundation, NVIDIA, NSF IIS 1651843, IIS 1700696. We also acknowledge computing support from Amazon. CF was supported by an NSF graduate research fellowship.
\bibliography{citations}
\bibliographystyle{icml2019}

\input{appendixa.tex}

\end{document}

%% file: defs.tex
% table
\makeatletter
\def\adl@drawiv#1#2#3{%
        \hskip.5\tabcolsep
        \xleaders#3{#2.5\@tempdimb #1{1}#2.5\@tempdimb}%
                #2\z@ plus1fil minus1fil\relax
        \hskip.5\tabcolsep}
\newcommand{\cdashlinelr}[1]{%
  \noalign{\vskip\aboverulesep
           \global\let\@dashdrawstore\adl@draw
           \global\let\adl@draw\adl@drawiv}
  \cdashline{#1}
  \noalign{\global\let\adl@draw\@dashdrawstore
           \vskip\belowrulesep}}
\makeatother

\newcommand{\cmt}[1]{{\footnotesize\textcolor{red}{#1}}}
\newcommand{\todo}[1]{\cmt{(TODO: #1)}}

\newcommand{\eg}{e.g.\ }
\newcommand{\ie}{i.e.\ }
\newcommand{\etal}{et al.\ }

% Math operators and functions
\newcommand{\E}[2]{\operatorname{\mathbb{E}}_{#1}\left[#2\right]}
\newcommand{\EEE}{\mathbb{E}}
\newcommand{\density}{p}
\newcommand{\proposal}{q}  % proposal distribution
\newcommand{\target}{p}  % target distribution
\newcommand{\prop}{P}
\newcommand{\kl}[2]{\mathrm{D_{KL}}\left(#1\;\middle\|\;#2\right)}
\newcommand{\entropy}{\mathcal{H}}
\newcommand{\ent}{\mathcal{H}}
\newcommand{\sdots}{\,\cdot\,}
\newcommand{\func}{\mathbf{f}}

% Constant matrices and vectors.
\newcommand{\ones}{\boldsymbol{1}}
\newcommand{\eye}{\boldsymbol{I}}
\newcommand{\zeros}{\boldsymbol{0}}

%meta learning
\newcommand{\task}{\mathcal{T}}
\newcommand{\metatrain}{\mathcal{T}^\text{meta-train}}
\newcommand{\metatest}{\mathcal{T}^\text{meta-test}}

\newcommand{\metatrainset}{\{\mathcal{T}_i~;~ i=1..N\}}
\newcommand{\metatestset}{\{\mathcal{T}_j~;~ j=1..M\} }
%\newcommand{\metatestset}{\{\mathcal{T}_j~;~ j\!\!=\!\!1..M\}^\text{meta-test} }

% parameters
\newcommand{\vphi}{\boldsymbol{\phi}}
\newcommand{\vtheta}{\boldsymbol{\theta}}
\newcommand{\vmu}{\boldsymbol{\mu}}

% supervised
\newcommand{\vx}{\mathbf{x}}
\newcommand{\vy}{\mathbf{y}}

% MDP
\newcommand{\sspace}{\mathcal{S}}
\newcommand{\aspace}{\mathcal{A}}
\newcommand{\state}{\mathbf{s}}
\newcommand{\sz}{{\state_0}}
\newcommand{\stm}{{\state_{t-1}}}
\newcommand{\st}{{\state_t}}
\newcommand{\sti}{{\state_t^{(i)}}}
\newcommand{\sT}{{\state_T}}
\newcommand{\stp}{{\state_{t+1}}}
\newcommand{\stpi}{{\state_{t+1}^{(i)}}}
\newcommand{\pdyn}{\density_\state}
\newcommand{\horizon}{T}
\newcommand{\hm}{{T-1}}
\newcommand{\action}{\mathbf{a}}
\newcommand{\az}{{\action_0}}
\newcommand{\atm}{{\action_{t-1}}}
\newcommand{\at}{{\action_t}}
\newcommand{\ati}{{\action_t^{(i)}}}
\newcommand{\atj}{{\action_t^{(j)}}}
\newcommand{\attildej}{{\tilde{\action}_t^{(j)}}}
\newcommand{\atij}{{\action_t^{(i,j)}}}
\newcommand{\atp}{{\action_{t+1}}}
\newcommand{\aT}{{\action_T}}
\newcommand{\atk}{{\action_t^{(k)}}}
\newcommand{\aTm}{\action_{\horizon-1}}
\newcommand{\opt}{^*}
% set of MDPs
\newcommand{\mdps}{\mathcal{M}}

% Trajectories
\newcommand{\demos}{\mathcal{D}}
\newcommand{\traj}{\tau}
\newcommand{\ptraj}{\density_\traj}
\newcommand{\visits}{\rho}  % Discounted visittation frequency

% Rewards
\newcommand{\reward}{r}
\newcommand{\rz}{\reward_0}
\newcommand{\rt}{\reward_t}
\newcommand{\rti}{\reward^{(i)}_t}
\newcommand{\rmi}{r_\mathrm{min}}
\newcommand{\rmax}{r_\mathrm{max}}
\newcommand{\return}{\eta}

% Loss
\newcommand{\loss}{\mathcal{L}}

% Optimality
\newcommand{\policyopt}{\pi^*}

% Value and Q function
\newcommand{\V}{V}
\newcommand{\Vsoft}{V_\mathrm{soft}}
\newcommand{\Vsoftparams}{V_\mathrm{soft}^\qparams}
\newcommand{\Vhatsoftparams}{\hat V_\mathrm{soft}^\qparams}
\newcommand{\Vhatsoft}{\hat V_\mathrm{soft}}
\newcommand{\Vhard}{V^{\dagger}}
\newcommand{\Q}{Q}
\newcommand{\Qsoft}{Q_\mathrm{soft}}
\newcommand{\Qsoftparams}{Q_\mathrm{soft}^\qparams}
\newcommand{\Qhatsoft}{\hat Q_\mathrm{soft}}
\newcommand{\Qhatsoftparams}{\hat Q_\mathrm{soft}^{\bar\qparams}}
\newcommand{\Qhard}{Q^{\dagger}}
\newcommand{\A}{A}
\newcommand{\Asoft}{A_\mathrm{soft}}
\newcommand{\Asoftparams}{A_\mathrm{soft}^\qparams}
\newcommand{\Ahatsoft}{\hat A_\mathrm{soft}}

% EBM
\newcommand{\energy}{\mathcal{E}}

\newcommand{\tasklosstrain}{\loss_{\task}^{\text{ tr}}}
\newcommand{\tasklosstest}{\loss_{\task}^{\text{ test}}}

\newcommand{\tasklosstraini}{\loss_{\task_i}^{\text{ tr}}}
\newcommand{\tasklosstesti}{\loss_{\task_i}^{\text{ test}}}

%% file: abstract.tex
\begin{abstract}
A significant challenge for the practical application of reinforcement learning to real world problems is the need to specify an oracle reward function that correctly defines a task. 
Inverse reinforcement learning (IRL) seeks to avoid this challenge by instead inferring a reward function from expert demonstrations.
While appealing, it can be impractically expensive to collect datasets of demonstrations that cover the variation common in the real world (e.g. opening any type of door).
Thus in practice, IRL must commonly be performed with only a limited set of demonstrations where it can be exceedingly difficult to unambiguously recover a reward function. In this work, we exploit the insight that demonstrations from other tasks can be used to constrain the set of possible reward functions by learning a ``prior'' that is specifically optimized for the ability to infer expressive reward functions from limited numbers of demonstrations. We demonstrate that our method can efficiently recover rewards from images for novel tasks and provide intuition as to how our approach is analogous to learning a prior.
\end{abstract}

%% file: intro.tex
\section{Introduction}
Reinforcement learning (RL) algorithms have the potential to automate a wide range of decision-making and control tasks across a variety of different domains, as demonstrated by successful recent applications ranging from robotic control~\citep{kober2012reinforcement,levine2016end} to game playing~\citep{mnih2015human,silver2016mastering}.
A key assumption of the RL problem statement is the availability of a reward function that accurately describes the desired task.
For many real world tasks, reward functions can be challenging to manually specify, while being crucial for good performance~\citep{amodei2016concrete}. 
Most real world tasks are multifaceted and require reasoning over multiple factors in a task (e.g. a robot cleaning in a house with children), while simultaneously providing appropriate reward shaping to make the task feasible with tractable exploration~\citep{ng1999policy}.
These challenges are compounded by the inherent difficulty of specifying rewards for tasks with high-dimensional observation spaces such as images.

\input{wrapfigs/front_fig.tex}

Inverse reinforcement learning (IRL) is an approach that aims to address this problem by instead inferring the reward function from demonstrations of the task \citep{Ng2000}.
This has the appealing benefit of taking a data-driven approach to reward specification in place of hand engineering.
In practice however, rewards functions are rarely learned as it can be prohibitively expensive to provide demonstrations that cover the variability common in real world tasks (e.g., collecting demonstrations of opening every type of door knob).
In addition, while learning a complex function from high dimensional observations might make an expressive function approximator seem like a reasonable modelling assumption, in the ``few-shot'' domain it is notoriously difficult to unambiguously recover a good reward function with expressive function approximators.
Many prior approaches have thus relied on low-dimensional linear models with handcrafted features that effectively encode a strong prior on the relevant features of a task.
This requires engineering a set of features by hand that work well for a specific problem.
In this work, we propose an approach that instead explicitly learns expressive features that are robust even when learning with limited demonstrations.

Our approach relies on the key observation that related tasks share a common structure that we can leverage when learning new tasks. %``prior''.
To illustrate, considering a robot navigating through a home. 
While the exact reward function we provide to the robot may differ depending on the task, there is a structure amid the space of useful behaviours, such as navigating to a series of landmarks, and there are \emph{certain behaviors} we always want to encourage or discourage, such as avoiding obstacles or staying a reasonable distance from humans.
This notion agrees with our understanding of why humans can easily infer the intents and goals (i.e., reward functions) of even abstract agents from just one or a few demonstrations~\cite{baker2007goal}, as humans have access to strong priors about how other humans accomplish similar tasks accrued over many years.
Similarly, our objective is to discover the common structure among different tasks, and encode that structure in a way that can be used to infer reward functions from  a few demonstrations.

More specifically, in this work we assume access to a set of tasks, along with demonstrations of the desired behaviors for those tasks, which we refer to as the \emph{meta-training set}.
From these tasks, we then learn a reward function parameterization that enables effective few-shot learning when used to initialize IRL in a novel task.
Our method is summarized in Fig.~\ref{fig:front_fig}.
Our key contribution is an algorithm that enables efficient learning of new reward functions by using meta-training to build a rich ``prior'' for goal inference.
Using our proposed approach, we show that we can learn deep neural network reward functions from raw pixel observations on two distinct domains with substantially better data efficiency than existing methods and standard baselines.

%% file: wrapfigs/front_fig.tex
\begin{figure}
    \begin{center}
    \includegraphics[width=\columnwidth]{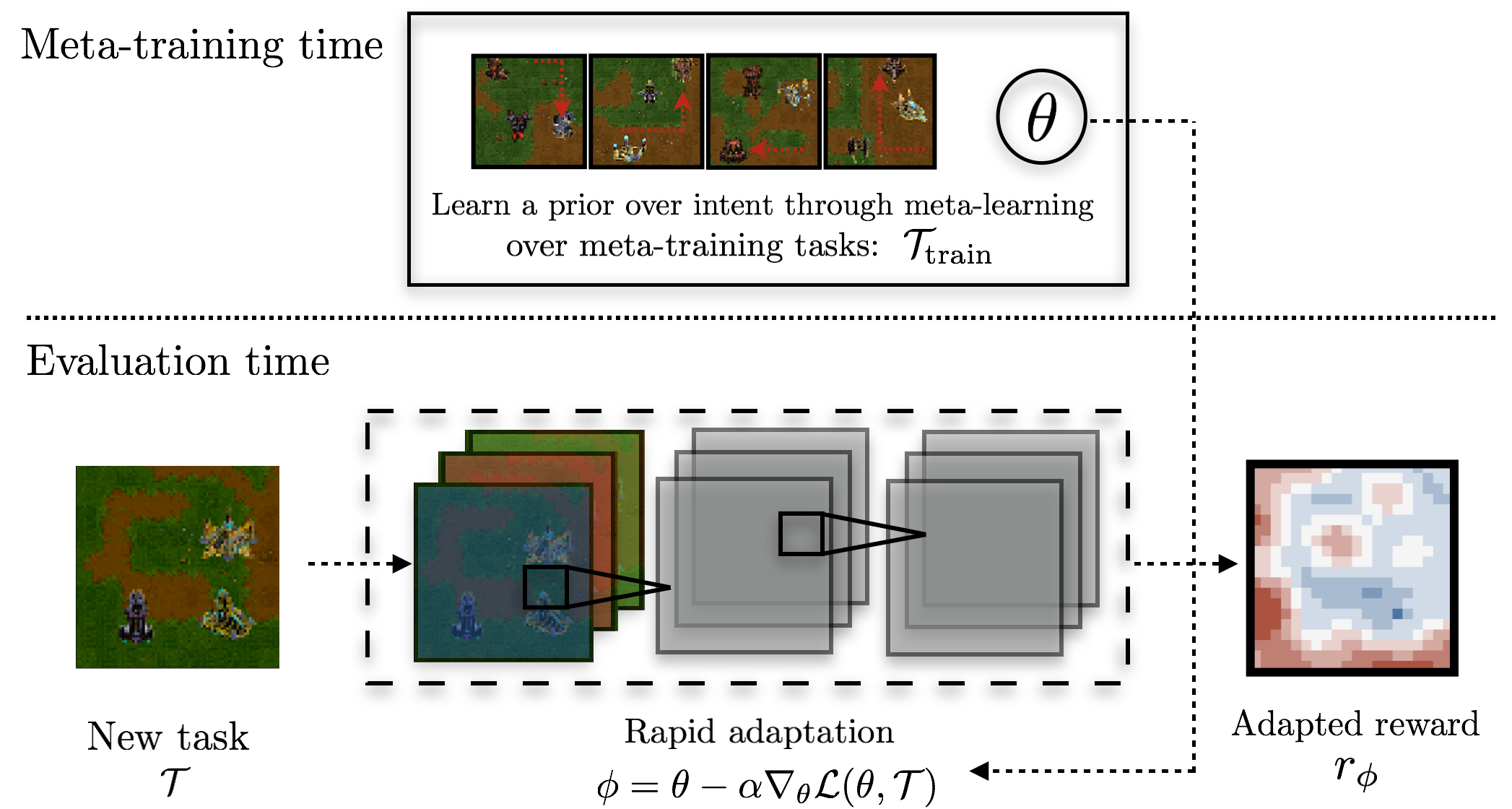}
    \end{center}
    %\vspace{-.1in}
    \captionsetup{font=footnotesize}
    \caption{A diagram of our meta-inverse RL approach. Our approach attempts to remedy over-fitting in few-shot IRL by learning a ``prior'' that constraints the set of possible reward functions to lie within a few steps of gradient descent. Standard IRL attempts to recover the reward function directly from the available demonstrations. The shortcoming of this approach is that there is little reason to expect generalization as it is analogous to training a density model with only a few examples.
     }
    \label{fig:front_fig}
    \vspace{-.255in}
\end{figure}

%% file: related_work.tex
\vspace{-2mm}
\section{Related Work}
\vspace{-1mm}
Inverse reinforcement learning (IRL) \citep{Ng2000} is the problem of inferring an expert's reward function directly from demonstrations.
Prior methods for performing IRL range from margin based approaches \citep{Abbeel2004, ratliff2006maximum} to probabilistic approaches \citep{Ramachandran2007,Ziebart2008}.
Although it is possible to extend our approach to any other IRL method, in this work we base on work on the maximum entropy (MaxEnt) framework~\citep{Ziebart2008}.
In addition to allowing for sub-optimality in the expert demonstrations, MaxEnt-IRL can be re-framed as a maximum likelihood estimation problem. (Sec.~\ref{sec:prelim}).

In part to combat the under-specified nature of IRL, prior work has often used low-dimensional linear parameterizations with handcrafted features~\citep{Abbeel2004,Ziebart2008}.
In order to learn from high dimensional input, \citet{Wulfmeier2015a} 
proposed applying fully convolutional networks~\citep{Shelhamer2017} to the MaxEnt IRL framework~\citep{Ziebart2008} for several navigation tasks~\citep{Wulfmeier2015b,Wulfmeier2016}.
Other methods that have incorporated neural network rewards
include guided cost learning (GCL) \citep{Finn2017}, which uses importance sampling and regularization for scalability to high-dimensional spaces, and adversarial IRL \citep{fu2018learning}.
Several other methods have also proposed imitation learning approaches based on adversarial frameworks that resemble IRL, but do not aim to directly recover a reward function~\citep{Ho2016,li2017inferring,hausman2017multi,Kuefler2018}.
In this work, instead of improving the ability to learn reward functions on a single task, we focus on the problem of effectively learning to use prior demonstration data from other IRL tasks, allowing us to learn new tasks from a limited number demonstrations even with expressive non-linear reward functions.

Prior work has explored the problem of \emph{multi-task} IRL, where the demonstrated behavior is assumed to have originated from multiple experts achieving different goals. Some of these approaches include those that aim to incorporate a shared prior over reward functions through extending the Bayesian IRL~\citep{Ramachandran2007} framework to the multi-task setting~\citep{Dimitrakakis2011, Choi2012}. Other approaches have clustered demonstrations while simultaneously inferring reward functions for each cluster~\citep{BabesVroman2011} or introduced regularization between rewards to a common ``shared reward''~\citep{Li2017}.
Our work is similar in that we also seek to encode prior information common to the tasks. 
However, a critical difference is that our method specifically aims to distill the meta-training tasks into a prior that can then be used to learn rewards for \emph{new} tasks efficiently.
The goal therefore is not to acquire good reward functions that explain the meta-training tasks, but rather to use them to learn efficiently on new tasks.

Our approach builds on work on the broader problem of meta-learning~\citep{schmidhuber1987evolutionary, bengio1990learning, naik1992meta, thrun2012learning} and generative modelling~\citep{rezende2016one, reed2018fewshot, mordatch2018}.
Prior work has proposed a variety of solutions for learning to learn including memory based methods \citep{duan2016rl, santoro2016meta, wang2016learning, mishra2017meta}, methods that learn an optimizer and/or initialization \citep{andrychowicz2016learning,ravi2016optimization,Finn2017,li2017learning}, and methods that compare new datapoints in a learned metric space \citep{koch2015siamese, Wang-2016-4848,vinyals2016matching, shyam2017attentive, snell2017prototypical}.
Our work is motivated by the goal of broadening the applicability of IRL, but in principle it is possible to adapt many of these meta-learning approaches for our problem statement. We build upon \citet{Finn2017}, which has also been previously applied to the related problems of imitation learning and human motion prediction~\cite{wang2016learning, finn2017one, alet2018modular}. We leave it to future work to do a comprehensive investigation of different meta-learning approaches which could broaden the applicability of IRL.

%% file: prelim.tex
\vspace{-2mm}
\section{Preliminaries and Overview}
\label{sec:prelim}
\vspace{-1mm}

In this section, we introduce our notation and describe the IRL and meta-learning problems.

\vspace{-2mm}
\subsection{Learning Rewards via Maximum Entropy Inverse Reinforcement Learning}\label{sec:maxent}
\vspace{-1mm}

The standard Markov decision process (MDP) is defined by the tuple $( \sspace, \aspace, \pdyn, \reward, \gamma )$ 
where $\sspace$ and $\aspace$ denote the set of possible states and actions respectively, $\reward:\sspace \times \aspace \rightarrow \mathbb{R}$ is the reward function, $\gamma\in[0,1]$ is the discount factor and $\pdyn:\  \sspace \times \sspace \times \aspace \rightarrow [0,\, 1]$ denotes the transition distribution over the next state $\stp$, given the current state $\st$ and current action $\at$. Typically, the goal of ``forward'' RL is to maximize the expected discounted return $R(\tau) = \sum_{t=1}^{T} \gamma^{t-1} r(\st, \at)$.

In IRL, we instead assume that the reward function is unknown but that we instead have access to a set of expert demonstrations $\demos = \{ \tau_1, \dots, \tau_K\}$, where $\tau_k = \{\state_1, \action_1, \dots, \sT, \aT\}$.

The goal of IRL is to recover the unknown reward function $\reward$ from the set of demonstrations.
We build on the maximum entropy (MaxEnt) IRL framework by ~\citet{Ziebart2008}, which models trajectories as being distributed proportional to their exponentiated return
\begin{equation}
    \label{eq:p_traj}
    \density(\tau) = \frac{1}{Z} \exp\left(R(\tau)\right),
\end{equation}
where $Z$ is the partition function, $Z = \int_\tau \exp(R(\tau)) \text{d}\tau$. This distribution can be shown to be induced by the optimal policy in entropy regularized forward RL problem:
\begin{equation}
    \label{eq:forward_rl}
    \policyopt = \argmax_{\pi} \E{\tau \sim \pi}{R(\tau) - \log\pi(\tau)}.
\end{equation}
This formulation allows us to pose the reward learning problem as a maximum likelihood estimation (MLE) problem in an energy-based model $\reward_{\vphi}$ by defining the following loss:
\begin{equation}
    \label{eq:inverse_rl}
    \min_{\vphi} \E{\tau \sim \demos}{\mathcal{L}_{\text{IRL}}(\tau)}\! :=\! \min_{\vphi} \E{\tau \sim \demos}{-\log \density_{\vphi} (\tau)}.
\end{equation}
Learning in general energy-based models of this form is common in many applications such as structured prediction.
However, in contrast to applications  where learning can be supervised by millions of labels (e.g. semantic segmentation), the learning problem in Eq.~\ref{eq:inverse_rl} must typically be performed with a relatively small number of example demonstrations.
In this work, we seek to address this issue in IRL by providing a way to integrate information from prior tasks to constrain the optimization in Eq. \ref{eq:inverse_rl} in the regime of limited demonstrations.

\vspace{-2mm}
\subsection{Meta-Learning}
\vspace{-1mm}
\label{sec:meta_learning_intro}
The goal of meta-learning algorithms is to optimize for the ability to learn efficiently on new tasks.
Rather than attempting to generalize to new datapoints, meta-learning can be understood as attempting to generalize to \textit{new tasks}.
It is assumed in the meta-learning setting that there are two \textit{disjoint} sets of tasks that we refer to as the meta-training set $\metatrainset$ and meta-test set $\metatestset$, which are both drawn from a distribution $p(\task)$.
During meta-training time, the meta-learner attempts to learn the structure of the tasks in the meta-training set, such that when it is presented with a test task, it can leverage this structure to learn efficiently from a limited number of examples. 

To illustrate this distinction, consider the case of few-shot learning setting.
Let $f_{\vtheta}$ denote the learner, and let a task be defined by learning from $K$ training examples $X_{\task}^{\text{tr}} = \{\vx_1 \dots, \vx_K\}$, $Y_{\task}^{\text{tr}} = \{\vy_1 \dots, \vy_K\}$, and evaluating on $K'$ test examples $X_{\task}^{\text{test}} = \{\vx_1 \dots, \vx_{K'}\}$, $Y_{\task}^{\text{test}} = \{\vy_1 \dots, \vy_{K'}\}$. 
One approach to meta-learning is to directly parameterize the meta-learner with an expressive model such as a recurrent or recursive neural network \citep{duan2016rl,mishra2017meta} conditioned on the task training data and the inputs for the test task: $f_{\vtheta}(Y | X_\task^{\text{test}}, X_\task^{\text{tr}}, Y_\task^{\text{tr}})$. Such a model is optimized using log-likelihood across all tasks.
In this approach to meta-learning, since neural networks are known to be universal function approximators \citep{siegelmann1995computational}, any desired structure between tasks can be implicitly encoded.

Rather than learn a single black-box function, another approach to meta-learning is to learn components of the learning procedure such as the initialization~\citep{Finn2017} or the optimization algorithm~\citep{ravi2016optimization,andrychowicz2016learning}. In this work we extend the approach of model agnostic meta-learning (MAML) introduced by~\citet{Finn2017}, which learns an initialization that is adapted by gradient descent. 
Concretely, in the supervised learning case, given a loss function $\mathcal{L}(\vtheta, X_{\task}, Y_{\task})$ (e.g. cross-entropy), 
MAML performs the following optimization
\begin{align}
\label{eq:maml}
&\min_{\vtheta} \sum_{\task} \loss(\vphi_{\task}, X_{\task}^{\text{test}}, Y_\task^{\text{test}})\nonumber \\
= &\min_{\vtheta} \sum_{\task} \loss \left(\vtheta - \alpha \nabla_{\vtheta} \loss (\vtheta, X_\task^{\text{tr}},  Y_\task^{\text{tr}}), X_\task^{\text{test}}, Y_\task^{\text{test}} \right),
\end{align}
where the optimization is over an initial set of parameters $\vtheta$ and the loss on the held out tasks $X_\task^{test}$ becomes the signal for learning the initial parameters for gradient descent (with step size $\alpha$) on $X_\task^{tr}$. This optimization is analogous to adding a constraint in a multi-task setting, which we show in later sections is analogous in our setting to learning a prior over reward functions.

%% file: methods.tex
\vspace{-2mm}
\section{Learning to Learn Rewards}
\label{sec:methods}
Our goal in meta-IRL is to learn how to learn reward functions across many tasks such that the model can infer the reward function for a new task using only one or a few expert demonstrations. Intuitively, we can view this problem as aiming to learn a prior over the rewards of expert demonstrators, such that when given just one or a few demonstrations of a new task, we can combine the learned prior with the new data to effectively determine the expert's intent. Such a prior is helpful in inverse reinforcement learning settings, since the space of reward functions with are relevant to particular task is much smaller than the space of all possible rewards definable on the raw observations.

During meta-training, we have a set of tasks $\metatrainset$. Each task $\task_i$ has a set of demonstrations \mbox{$\demos_\task = \{\tau_1, \dots, \tau_K\}$} from an expert policy which we partition into disjoint $\demos_\task^{\text{tr}}$ and $\demos_\task^{\text{test}}$ sets. The demonstrations for each meta-training task are assumed to be produced by the expert according to the maximum entropy model in Section~\ref{sec:maxent}. During meta-training, these tasks will be used to encodes common structure so that our model can quickly acquire rewards for new tasks from just a few demonstrations.

After meta-training, our method is presented with a new task. 
During this meta-test phase, the algorithm must infer the parameters of the reward function $r_{\vphi}(\st, \at)$ for the new task from a few demonstrations.
As is standard in meta-learning, we assume that the test task is from the same distribution of tasks seen during meta-training, a distribution that we denote as $p(\task)$.

\subsection{Meta Reward and Intention Learning (MandRIL)}

In order to meta-learn a reward function that can act as a prior for new tasks and new environments, we first formalize the notion of a good reward by defining a loss $\mathcal{L}_{\task}(\vtheta)$ on the reward function $r_{\vtheta}$ for a particular task $\task$. We use the MaxEnt IRL loss $\mathcal{L}_\text{IRL}$ discussed in Section~\ref{sec:prelim}, which, for a given $\demos_\task$, leads to the following gradient~\citep{Ziebart2008}: 
\begin{align}
\label{eq:irl_loss_grad}
\nabla_{\vtheta} \mathcal{L}_{\task}(\vtheta) & = \frac{\partial r_{\vtheta}}{\partial \vtheta} \,\left[ \mathbb{E}_{\tau}[\vmu_{\tau}] - \vmu_{\mathcal{D}_\task} \right]. %\notag
\end{align}
where $\vmu_\tau$ are the state-action visitations under the optimal maximum entropy policy under $r_{\vtheta}$, and $\vmu_{\mathcal{D}}$ are the mean state visitations under the demonstrated trajectories.

\input{pseudocode.tex}
If our end goal were to achieve a single reward function that works as well as possible across all tasks in $\metatrainset$, then we could simply follow the \emph{mean} gradient across all tasks. However, our objective is different: instead of optimizing performance on the meta-training tasks, we aim to learn a reward function that can be quickly and efficiently adapted to new tasks at meta-test time. In doing so, we aim to encode prior information over the task distribution in this learned reward prior.

We propose to implement such a learning algorithm by finding the parameters $\vtheta$,
such that starting from $\vtheta$ and taking a small number of gradient steps on a few demonstrations from given task leads to a reward function for which a set of \emph{test} demonstrations have high likelihood, with respect to the MaxEnt IRL model.
In particular, we would like to find a $\vtheta$ such that the parameters 
\begin{align}
\vphi_{\task} = \vtheta - \alpha \nabla_{\vtheta} \tasklosstrain(\vtheta)
\label{eq:irl_adapt}
\end{align}
lead to a reward function $r_{\vphi_{\task}}$ for task $\task$, such that the IRL loss (corresponding to negative log-likelihood) for a disjoint set of test demonstrations, given by $\loss_\text{IRL}^{\task,\text{test}}$, is minimized. The corresponding optimization problem for $\vtheta$ can therefore be written as follows:
\vspace{-0.15cm}
\begin{align}
\label{eq:maml_irl}
\displaystyle \min_{\vtheta} \sum_{i=1}^N \tasklosstesti(\vphi_{\task_i}) =
\sum_{i=1}^N \tasklosstesti \left(\vtheta - \alpha \nabla_{\vtheta} \tasklosstraini(\vtheta)\right).
\vspace{-0.3cm}
\end{align}
Our method acquires this prior $\vtheta$ over rewards in the task distribution $p(\task)$ by optimizing this loss. This amounts to an extension of the MAML algorithm in Section~\ref{sec:meta_learning_intro} to the inverse reinforcement learning setting. This extension is quite challenging, because computing the MaxEnt IRL gradient requires repeatedly solving for the current maximum entropy policy and visitation frequencies, and the MAML objective requires computing derivatives \emph{through} this gradient step. Next, we describe in detail how this is done.
An overview of our method is also outlined in Alg.~\ref{alg:meta_training}.

\noindent\textbf{Meta-training. }
The computation of the meta-gradient for the objective in Eq.~\ref{eq:maml_irl} can be conceptually separated into two parts. First, we perform the update in Eq.~\ref{eq:irl_adapt} by computing the \emph{expected state visitations} $\vmu$, which is the expected number of times an agent will visit each state. We denote this overall procedure as \textsc{State-Visitations-Policy}, and follow \citet{Ziebart2008} by first computing the maximum entropy optimal policy in Eq.~\ref{eq:forward_rl} under the current $r_{\vtheta}$, and then approximating $\vmu$ using dynamic programming. Next, we compute the state visitation distribution of the expert using a procedure which we denote as \textsc{State-Visitations-Traj}. This can be done either empirically, by averaging the state visitation of the experts demonstrations, or by using \textsc{State-Visitations-Policy} if the true reward is available at meta-training time. This allows us to recover the IRL gradient according to Eq.~\ref{eq:irl_loss_grad}, which we can then apply to compute $\vphi_\task$ according to Eq.~\ref{eq:irl_adapt}.

Second, we need to differentiate through this update to compute the gradient of the meta-loss in Eq.~\ref{eq:maml_irl}. Note that the meta-loss itself is the IRL loss evaluated with a different set of test demonstrations. We follow the same procedure as above to evaluate the gradient of $\loss_\text{IRL}^{\task,\text{test}}$ with respect to the post-update parameters $\vphi_\task$, and then apply the chain rule to compute the meta-gradient:
\begin{align}
    \label{eq:grad_meta_obj}
    \nabla_\theta \tasklosstest (\vtheta)
    = \frac{\partial }{\partial\vtheta}(\vtheta - \alpha \nabla_{\vtheta} \tasklosstrain(\vtheta)) \frac{\partial r_{\vphi_\task}}{\partial\vphi_\task} \frac{\partial\tasklosstest}{\partial r_{\vphi_\task}}& \nonumber \\
    = \left(\mathbf{I} - \alpha \frac{\partial^2 \tasklosstrain(\vtheta)}{\partial \vtheta^2} - \alpha \frac{\partial\,r_{\vtheta}}{\partial\,\vtheta} D \frac{\partial\,r_{\vtheta}}{\partial\,\vtheta}^\top \right)  \frac{\partial r_{\vphi_\task}}{\partial\vphi_\task} \frac{\partial\tasklosstest}{\partial r_{\vphi_\task}}&
\end{align}

where on the second line we differentiate through the MaxEnt-IRL update, and we define the $|\mathcal{S}| |\mathcal{A}|$-dimensional diagonal matrix $D$ as \[ D := \text{diag}\left(\left\{\frac{\partial\,}{\partial\,r_{\vtheta, i}}(\mathbb{E}_\tau [\mu_\tau])_i\right\}_{i=1}^{|\mathcal{S}| |\mathcal{A}|}\right). \]
A detailed derivation of this expression is provided in the supplementary Appendix~\ref{appendix:derivation}.

\noindent\textbf{Meta-testing. }
Once we have acquired the meta-trained parameters $\vtheta$ that encode a prior over $p(\task)$, we can leverage this prior to enable fast, few-shot IRL of novel tasks in $\metatestset$. For each task, we first compute the state visitations from the available set of demonstrations for that task. Next, we use these state visitations to compute the gradient, which is the same as the inner loss gradient computation of the meta-training loop in Alg.~\ref{alg:meta_training}. We apply this gradient to adapt the parameters $\vtheta$ to the new task. Even if the model was trained with only one inner gradient steps, we found in practice that it was beneficial to take substantially more gradient steps during meta-testing; performance continued to improve with up to 20 steps.

\vspace{-0.3cm}
\subsection{Connection to Learning a Prior over Intent}
\label{sec:prior}

The objective in Eq.~\ref{eq:irl_adapt} optimizes for parameters that enable the reward function to generalize efficiently on a wide range of tasks. Intuitively, constraining the space of reward functions to lie within a few steps of gradient descent can be interpreted as expressing a ``locality'' prior over reward function parameters. This intuition can be made more concrete by the following analysis.
\input{wrapfigs/graphical_model.tex}

By viewing IRL as maximum likelihood estimation in a particular graphical model (Fig.~\ref{fig:graphical}), we can take the perspective of ~\citet{grant2017recasting} who showed that for a linear model, fast adaptation via a few steps of gradient descent in MAML is performing MAP inference over $\vphi$, under a Gaussian prior with the mean $\vtheta$ and a covariance that depends on the step size, number of steps and hessian of the loss.
This is based on the connection between early stopping and regularization previously discussed in~\citet{santos1996equivalence}, which we refer the readers to for a more detailed discussion.
The interpretation of MAML as imposing a Gaussian prior on the parameters is exact in the case of a likelihood that is quadratic in the parameters (such as the log-likelihood of a Gaussian in terms of its mean).
For any non-quadratic likelihood, this is an approximation in a local neighborhood around $\vtheta$ (i.e. up to convex quadratic approximation). In the case of complex parameterizations, such as deep function approximators, this is a coarse approximation and unlikely to be the mode of a posterior. However, we can still frame the effect of early stopping and initialization as serving as a prior in a similar way as prior work~\citep{sjoberg1995overtraining,duvenaud2016early,grant2017recasting}. More importantly, this interpretation hints at future extensions to our approach that could benefit from employing more fully Bayesian approaches to reward and goal inference.

%% file: pseudocode.tex
\begin{algorithm}[tb]
\caption{Meta Reward and Intention Learning (MandRIL)}
\label{alg:meta_training}
\begin{algorithmic}[1]
\STATE {\bfseries Input:} 
Set of meta-training tasks $\{\task\}^{\text{meta-train}}$
\STATE {\bfseries Input:} hyperparameters $\alpha, \beta$
\FUNCTION{\textsc{MaxEntIRL-Grad}($r_{\vtheta}$, $\task$, $\demos$)}{} %\COMMENT{Single task update}
    \STATE \# \textit{Compute state visitations of demos}
    \STATE $\vmu_{\mathcal{D}} = $ \textsc{State-Visitations-Traj}($\task$, $\demos$)
    \STATE \# \textit{Compute Max-Ent state visitations}
    \STATE $\mathbb{E}_{\tau}[\vmu_{\tau}] =$ \textsc{State-Visitations-Policy}($r_{\vtheta}$, $\task$)
    \STATE \# \textit{MaxEntIRL gradient} \citep{Ziebart2008}
    \STATE $\frac{\partial\loss}{\partial r_{\vtheta}} = \mathbb{E}_{\tau}[\vmu_{\tau}] - \vmu_{\mathcal{D}}$ 
    \STATE \textbf{Return} $\frac{\partial\loss}{\partial r_{\vtheta}}$
\ENDFUNCTION
\STATE
\STATE Randomly initialize $\vtheta$
\WHILE{not done}
\STATE Sample batch of tasks $\task_i \sim \{\task\}^{\text{meta-train}}$ 
\FOR{{\bfseries all} $\task_i$}
    \STATE Sample demos $\demos^{ \text{tr}} = \{\tau_1, \dots, \tau_{K} \} \sim \task_i$
    % \STATE Compute the reward function $r_i = r_\vtheta(\tau)$  
    \STATE \# \textit{Inner loss computation}
    \STATE $\frac{\partial\tasklosstraini(\vtheta)}{\partial r_{\vtheta}} =$ \textsc{MaxEntIRL-Grad}($r_{\vtheta}$, $\task_i$, $\demos^{\text{tr}}$)
    \STATE Compute $\nabla_{\vtheta} \tasklosstraini(\vtheta)$ from $\frac{\partial\tasklosstraini (\vtheta)}{\partial r_{\vtheta}}$
    \STATE Compute $\vphi_{\task_i} = \vtheta - \alpha\nabla_{\vtheta} \tasklosstraini (\vtheta)$
    \STATE Sample demos $\demos^{\text{test}} = \{\tau_1', \dots, \tau_{K'}' \} \sim \task_i$ 
    \STATE \# \textit{Outer loss computation}
    \STATE $\frac{\partial\tasklosstesti}{\partial r_{\vtheta}} =$
    \textsc{MaxEntIRL-Grad}($r_{\vphi_{\task_i}}$, $\task_i$, $\demos^{\text{test}}$)) 
    \STATE \# \textit{Compute meta-gradient}
    \STATE Compute $\nabla_{\vtheta} \tasklosstesti$ from $\frac{\partial\tasklosstesti}{\partial r_{\vtheta}}$ via chain rule 
\ENDFOR
\STATE Compute update to $\vtheta \leftarrow \vtheta - \beta \sum_i \nabla_{\vtheta} \tasklosstesti$ 
\ENDWHILE
\end{algorithmic}
\end{algorithm}

%% file: wrapfigs/graphical_model.tex
% For tikz
\newcommand*{\nodesep}{1.7cm}
\newcommand*{\facsep}{.35cm}
\newcommand*{\facsize}{.20cm}
\newcommand*{\rewfactortikz}{{\large $\mathbf{\Phi_{\text{r}}}$}}
\newcommand*{\dynfactortikz}{
{\large $\mathbf{\Phi_{\text{dyn}}}$}
}
\newcommand*{\rewfactor}{\mathbf{\Phi_{\text{r}}}}
\newcommand*{\dynfactor}{\mathbf{\Phi_{\text{dyn}}}}
%https://xkcd.com/color/rgb/
\definecolor{rew_col}{HTML}{000000}
\definecolor{dyn_col}{HTML}{000000}
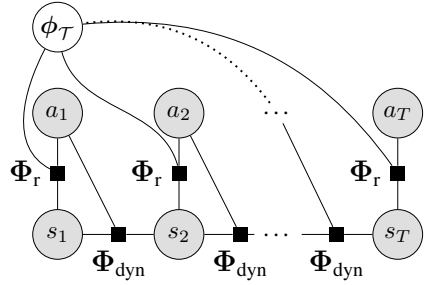
\begin{figure}
    \centering
    \resizebox{0.7\columnwidth}{!}{
        \begin{tikzpicture}[->,auto,node distance=1.4cm]
        \node[obs](A1){$a_1$};
        \node[obs, right= of A1](A2){$a_2$};
        \node[obs, below= of A1](S1){$s_1$};
        \node[obs, below= of A2](S2){$s_2$};
        \node(DOTS)[right of=S2]{\ldots};
        \node(DOTA)[right of=A2]{\ldots};
        \node[obs, right= of DOTS](ST){$s_T$};
        \node[obs, above= of ST](AT){$a_T$};
        \factor[above=of S1,fill=rew_col, minimum size=\facsize] {P1} {left:\rewfactortikz} {} {}
        \factoredge{A1, S1} {P1} {}; 
        \factor[right=of S1,fill=dyn_col, minimum size=\facsize] {D1} {below:\dynfactortikz} {} {};
        \factoredge {A1, S1, S2} {D1} {} ;
        \factor[above=of S2,fill=rew_col, minimum size=\facsize] {P2} {left:\rewfactortikz} {} {}
        \factoredge{A2, S2} {P2} {}; 
        \factor[right=of S2,fill=dyn_col, minimum size=\facsize] {D1} {below:\dynfactortikz} {} {};
        \factoredge {A2, S2, DOTS} {D1} {} ;
        \factor[above=of ST,fill=rew_col, minimum size=\facsize] {PT} {left:\rewfactortikz} {} {}
        \factoredge{AT, ST} {PT} {}; 
        \factor[left=of ST,fill=dyn_col, minimum size=\facsize] {PTM1} {below:\dynfactortikz} {} {}
        \factoredge{DOTS,ST,DOTA} {PTM1} {};
        \node[latent, above=0.5cm of A1, ](T){$\vphi_\task$};
        \factoredge[bend left, out=60]{T, P1}{T} {};
        \factoredge[bend left, out=-30]{T, P2}{T} {};
        \factoredge[bend right, thick, dotted]{T, DOTA}{T} {};
        \factoredge[bend right]{T, PT}{T} {};
        \end{tikzpicture}
    }
    \vspace{-0.3cm}
    \captionsetup{font=footnotesize}
    \caption{Our approach can be understood as approximately learning
            a distribution over the demonstrations $\tau$, in the factor graph $p(\tau) = \frac{1}{Z} \prod_{t=1}^{T}\rewfactor(\vphi_\task, \st, \at) \dynfactor(\stp, \st, \at)$ (above) where we learn a prior over $\phi_\task$, which during meta-test is used for MAP inference over new expert demonstrations.}
    \label{fig:graphical}
    \vspace{-.5cm}
\end{figure}

%% file: suncg_experiments.tex
\vspace{-0.3cm}
\section{Experiments}
\label{sec:experiments}

Our evaluation seeks to answer two questions. First, we aim to test our core hypothesis that using prior task experience enables reward learning for new tasks with just a few demonstrations. Second, we compare our method with alternative approaches that make use of multi-task experience.

We test our core hypothesis by comparing learning performance on a new tasks starting from the initialization produced by MandRIL with learning a separate model for every task starting either from a random initialization or from 
an initialization obtained by supervised pre-training. We refer to these approaches as learning \textsc{``from scratch''} and \textsc{``average gradient''} pretraining respectively. Our supervised pre-training baseline follows the average gradient during meta-training tasks and finetunes at meta-test time (as discussed in Section~\ref{sec:methods}). Unlike our method, supervised pre-training does not optimize for a model that performs well under fine tuning, but does use the same prior data to pre-train. We additionally compare to pre-training on a single task as well as all the meta-training tasks.

%This comparison evaluates whether our proposed method of learning an initializing can in fact make inverse RL more efficient. 

To our knowledge, there is no prior work that addresses the specific meta-inverse reinforcement learning problem introduced in this paper. Thus, to provide a point of comparison and calibrate the difficulty of the tasks, we adapt two alternative black-box meta-learning methods to the IRL setting. The comparisons to both of the black-box methods described below evaluate the importance of incorporating the IRL gradient into the meta-learning process, rather than learning the adaptation process entirely from scratch.

\vspace{-0.1in}\paragraph{Demo conditional model:} Our method implicitly conditions on the demonstrations through the gradient update. In principle, a conditional deep model with sufficient capacity could implicitly implement a similar learning rule. Thus, we consider a conditional model (often referred to as a ``contextual model''~\citep{finn2017one}), which receives the demonstration as an additional input. 
\vspace{-0.1in}\paragraph{Recurrent meta-learner:} We additionally compare to an RNN-based meta-learner~\citep{santoro2016meta,duan2017oneshot}. Specifically, we implement a conditional model by feeding both images and sequences of states visited by the demonstrations to an LSTM.

\input{wrapfigs/environment_figure.tex}
\input{wrapfigs/suncg.tex}
\input{wrapfigs/reward_function_visualization.tex}

We consider two environments: \textbf{(1)} an image-based navigation task with an aerial viewpoint, \textbf{(2)} a first-person navigation task in a simulated home environment with object interaction. We describe here the environments and evaluation protocol and provide detailed experimental settings and hyperparameters for both domains in Appendices~\ref{appendix:spriteworld} and ~\ref{appendix:suncg}.
\vspace{-.1in}
\paragraph{(1)\, SpriteWorld navigation domain.}
Since most prior IRL work (and multi-task IRL work) studied settings where linear reward function approximators suffice (i.e., low-dimensional state spaces and hand-designed features), we design an experiment that is significantly more challenging---that requires learning rewards on raw pixels.
%---while still exhibiting multi-task structure needed to test our core hypothesis.
We consider a navigation problem where we must learn a convolutional neural network that directly maps image pixels to rewards. We introduce a family of tasks called ``SpriteWorld.'' Some example tasks are shown in Fig.~\ref{fig:gridworlds}. 
Tasks involve navigating to goal objects while exhibiting preference over terrain types (e.g., the agent prefers to traverse dirt tiles over traversing grass tiles). At meta-test time, we provide one or a few demonstrations in a single training environment and evaluate the reward learned using these demonstrations in a new, test environment that contains the same objects as the training environment, but arranged differently. Evaluating in a new test environment is critical to measure that, after adapting to the training environment from a few demonstrations, the reward learned the correct visual cues, rather than simply memorizing the demonstration trajectory. 

\input{wrapfigs/plots.tex}

We generate unique tasks in this domain as follows. First, we randomly choose a set of three sprites from one hundred sprites from the original game (creating a total of 161,700 unique tasks). We randomly place these three sprites within a randomly generated terrain tiling; we designate one of the sprites to be the goal landmark of the navigation task. The other two objects are treated as obstacles for which the agent incurs a large negative reward for not avoiding. In each task, we optimize our model on a training world and generalization in a test world, as described below.
\vspace{-0.2cm}
\paragraph{(2)\, SUNCG navigation domain:}

In addition to the SpriteWorld domain, we evaluate our approach on a first person image-based navigation task in an indoor house environment where the agent must interact with objects. We use an environment built on top of the SUNCG dataset~\citep{song2016ssc} which has previously been used in the context of IRL~\cite{fu2019language} with language instructions. We follow a similar task setup as \citet{fu2019language}, although we omit the language instructions. In this domain, we consider tasks that can be categorized into two types.

\vspace{-0.2cm}\paragraph{Navigation (NAV):} In this task, the agent must navigate to a location in the house that corresponds either to a target object or location. For example, in the blue line of Fig.~\ref{fig:suncg_task_example}, the agent must navigate to the ``cup'' object.
\vspace{-0.3cm}\paragraph{Pick-and-place (PICK):} In this more difficult task, the agent moves an object between two locations. For example, in Fig~\ref{fig:suncg_task_example}, the agent must navigate to the ``cup'', perform a pick action and then navigate to the ``bedroom''.

\vspace{-0.2cm}
\paragraph{Evaluation protocol:}
We evaluate on held-out tasks that were unseen during meta-training. In the SpriteWorld domain, we consider two settings: (1) tasks involving new combinations and placements of sprites, with sprites that were present during meta-training, and (2) tasks with combinations of unseen sprites which we refer to as ``out of domain objects.''
For each task, we generate one environment (a set of sprite positions) along with demonstrations for adapting the reward, and generate a second environment (with new sprite positions) for evaluating the adapted reward.

In the SUNCG domain, we similarly evaluate on both novel combinations of objects and locations. We follow the evaluation protocol of \citet{fu2019language} and evaluate on ``TEST'' tasks which consist of tasks within the same houses as training, but with novel combinations of objects and locations. In addition, we evaluate on environments which consists of new houses not in the training set. We refer to these as ``UNSEEN-HOUSES.'' This evaluation adds complexity by testing the models ability to successfully infer rewards in an entirely new scene. In total, the dataset consists of 1413 tasks (716 PICK, 697 NAV). The meta-train set is composed of 1004 tasks, the ``TEST'' set contains 236 tasks, and the ``UNSEEN-HOUSES'' set contains 173 tasks.

\vspace{-0.15in} \paragraph{Evaluation Metrics.}
We measure performance using the expected value difference, which measures the sub-optimality of a policy learned under the learned reward; this is a performance metric used in prior IRL work~\citep{levine2011nonlinear, Wulfmeier2015a}. The metric is computed by taking the difference between the value of the optimal policy under the learned reward and the value of the optimal policy under the true reward. On the SUNCG domain, we follow \citet{fu2019language} and report the success rate of the optimal policy under the learned reward function. 

\vspace{-0.15in}
\paragraph{Results.} The results for SpriteWorld are shown in Fig.~\ref{fig:results}, which illustrate test performance with in-distribution and out-of-distribution sprites. Our approach, MandRIL, achieves consistently better performance in both settings. Most significantly, our approach performs well even with single-digit numbers of demonstrations. By comparison, alternative meta-learning methods generally overfit considerably, attaining good training performance (see Appendix.~\ref{appendix:extra_plot} for curves) but poor test performance. Learning the reward function from scratch is in fact the most competitive baseline -- as the number of demonstrations increases, simply training the reward function from scratch on the new task is the only method that matches the performance of MandRIL when provided $20$ or more demonstrations. With only a few demonstrations however, MandRIL has substantially lower value difference. It is worth noting the performance of MandRIL on the out of distribution test setting (Fig.~\ref{fig:results}, bottom):
although the evaluation is on new sprites, MandRIL is still able to adapt via gradient descent and exceed the performance all other methods.

\vspace{-.05cm}
In both domains, we perform a comparison to representations finetuned from a supervised pre-training phase in Fig.~\ref{fig:results} and Table~\ref{tab:suncg_results}. We compare against an approach that follows the mean gradient across the tasks at meta-training time and is fine-tuned at meta-test time which we find consistently leads to negative transfer. We conclude that fine tuning reward functions learned in this manner is not an effective way of using prior task information. In contrast, we find that our approach, which explicitly optimizes for initial weights for fine-tuning, robustly improves performance on all task types and test settings. By visualizing the value under the learned reward function (see Fig.~\ref{fig:reward_viz}), we see that even with a small number of gradient steps, the reward function can be effectively adapted to an unseen home layout. 
\input{suncg_table.tex}

\vspace{-.05cm}
Note that the SUNCG task is substantially more challenging, requiring the reward function to interpret first-person images. Indeed, the pretrained MaxEntIRL algorithm that does not use meta-learning exhibits \emph{negative} transfer, as illustrated by the lower performance of this method on all tasks as compared to the learning \textsc{``from scratch''} version, which learns each task entirely from random initialization. Training from scratch is a strong baseline here, because the method still sees every single first-person image in the house
-- 2257.7 images on average.
This provides sufficient variety to learn effective visual features in many cases. Nonetheless, our method (last row in Table 1) produces a substantial improvement, especially on the much harder ``PICK'' task, demonstrating that meta-learning can produce \emph{positive} transfer even when pre-training does not.

%% file: wrapfigs/environment_figure.tex
\newcommand{\envfigsize}{0.8\columnwidth}
%\begin{wrapfigure}[19]{R}{\envfigsize}
\begin{figure}
    \centering
     \includegraphics[width=\envfigsize]{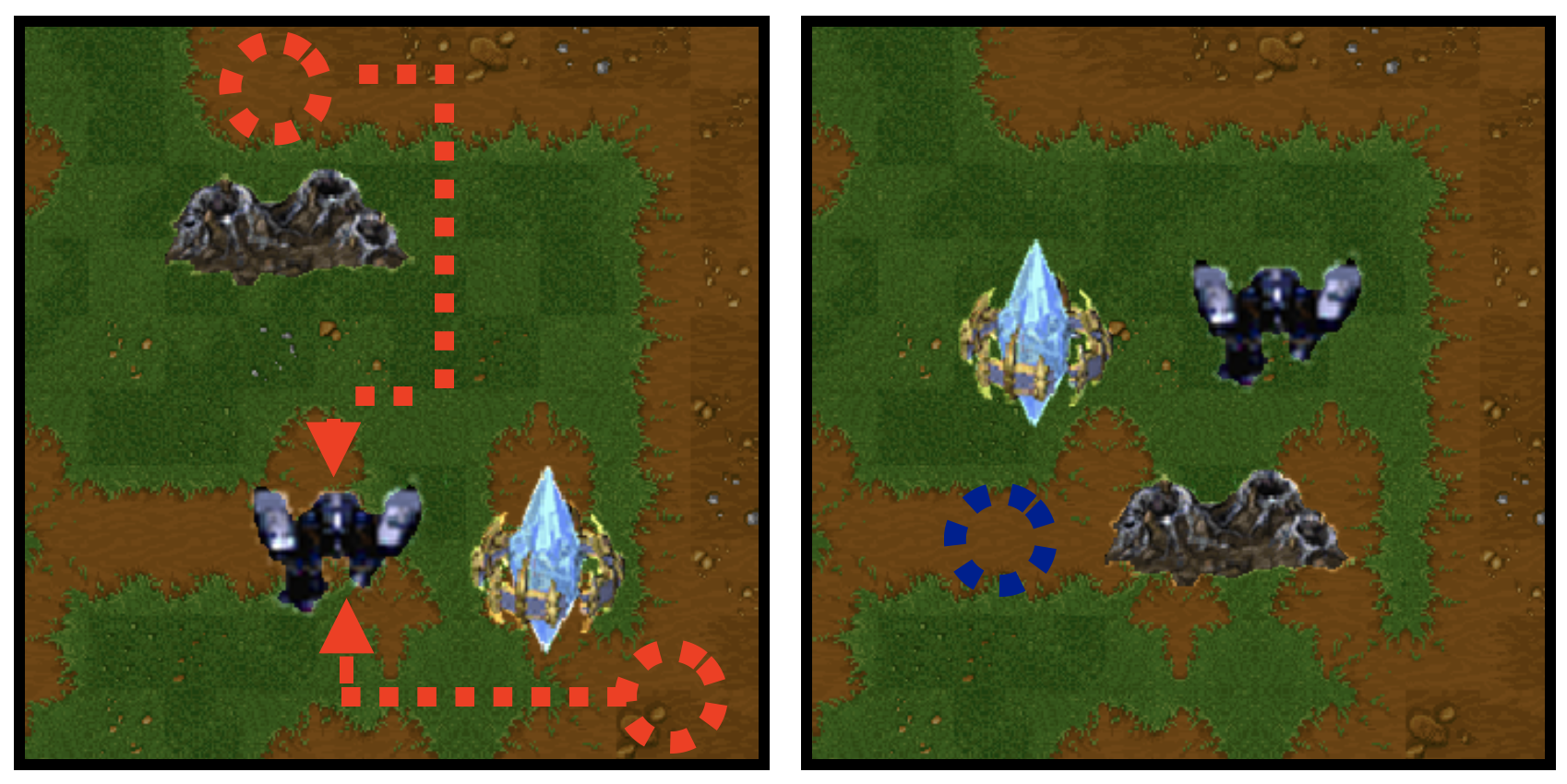} \\
     \vspace{-0.2cm}
    \captionsetup{font=footnotesize}
    \caption{\small An example task on the SpriteWorld domain. When learning a task, the agent has access to the image (left) and demonstrations (red arrows). To evaluate learning (right), the agent is tested for its ability to recover the reward for the task when the objects have been rearranged. The reward structure we wish to capture can be illustrated by considering the initial state in blue. An policy acting optimally under a correctly inferred reward should interpret the other objects as obstacles, and prefer a path on dirt.\label{fig:gridworlds}
    }
    \vspace{-0.2in}
\end{figure}
%\vspace{-6.0cm}

%% file: wrapfigs/suncg.tex
%\begin{wrapfigure}[19]{R}{\envfigsize}
\begin{figure}[ht]
    \centering
    \includegraphics[width=.8\linewidth]{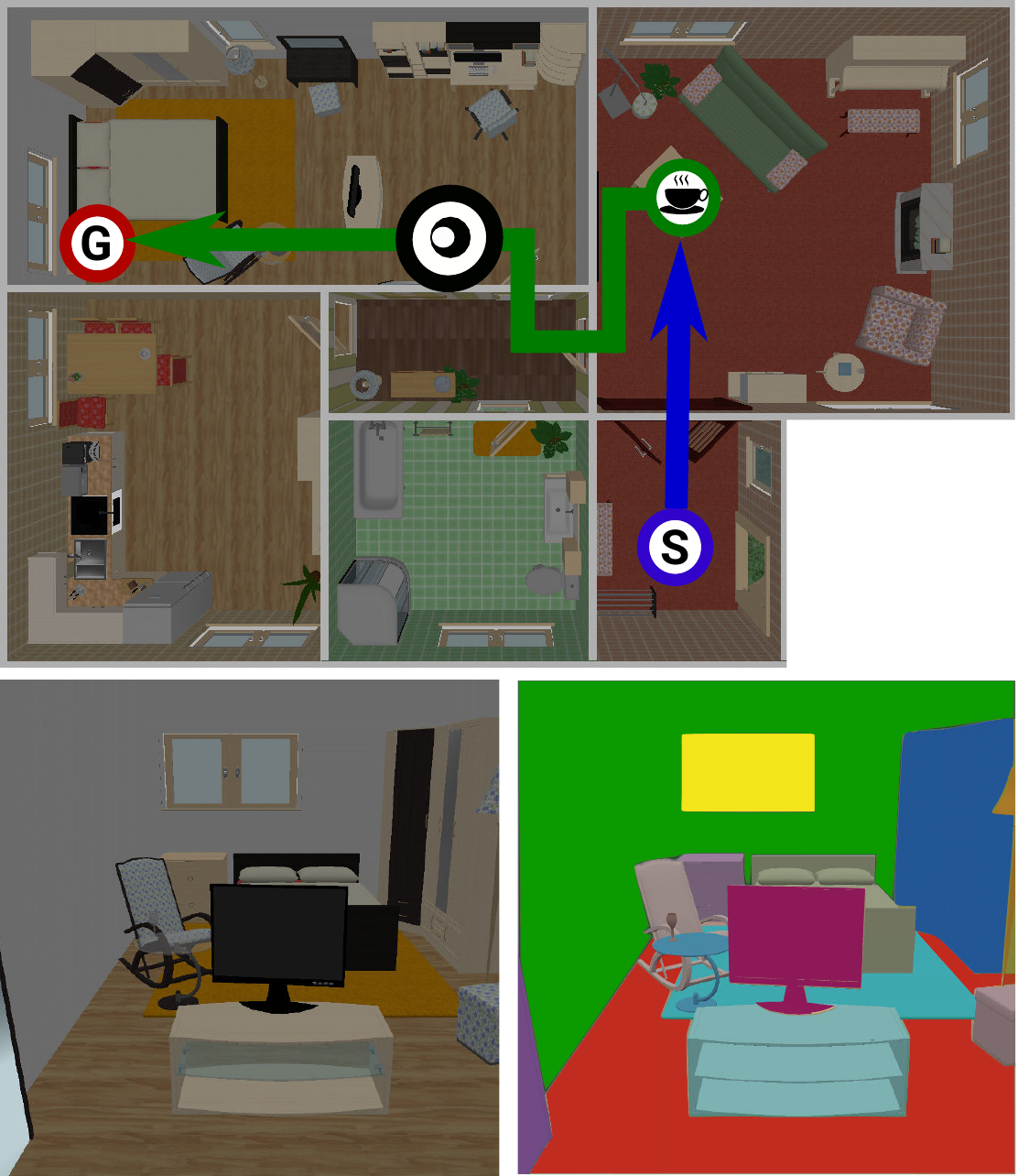}
    \vspace{-.25cm}
    \captionsetup{font=footnotesize}
    \caption{\small An example task in the SUNCG environment (top). The agent must complete either a ``NAV'' task (blue line) where the goal is to navigate from the start to the cup or a ``PICK'' task (blue + green line) where the agent must also bring the cup to the bed. The agent's observation is a panoramic first-person viewpoint (see bottom left for RGB). Following the convention in prior work~\cite{fu2019language}, we provide to the reward function the corresponding semantic images (bottom right). These images are $32\times24$ containing 61 channels corresponding to each object class.}
    \label{fig:suncg_task_example}
    \vspace{-0.5cm}
\end{figure}

%% file: wrapfigs/reward_function_visualization.tex
\begin{figure}[h]
    \centering
    \includegraphics[width=0.92\columnwidth]{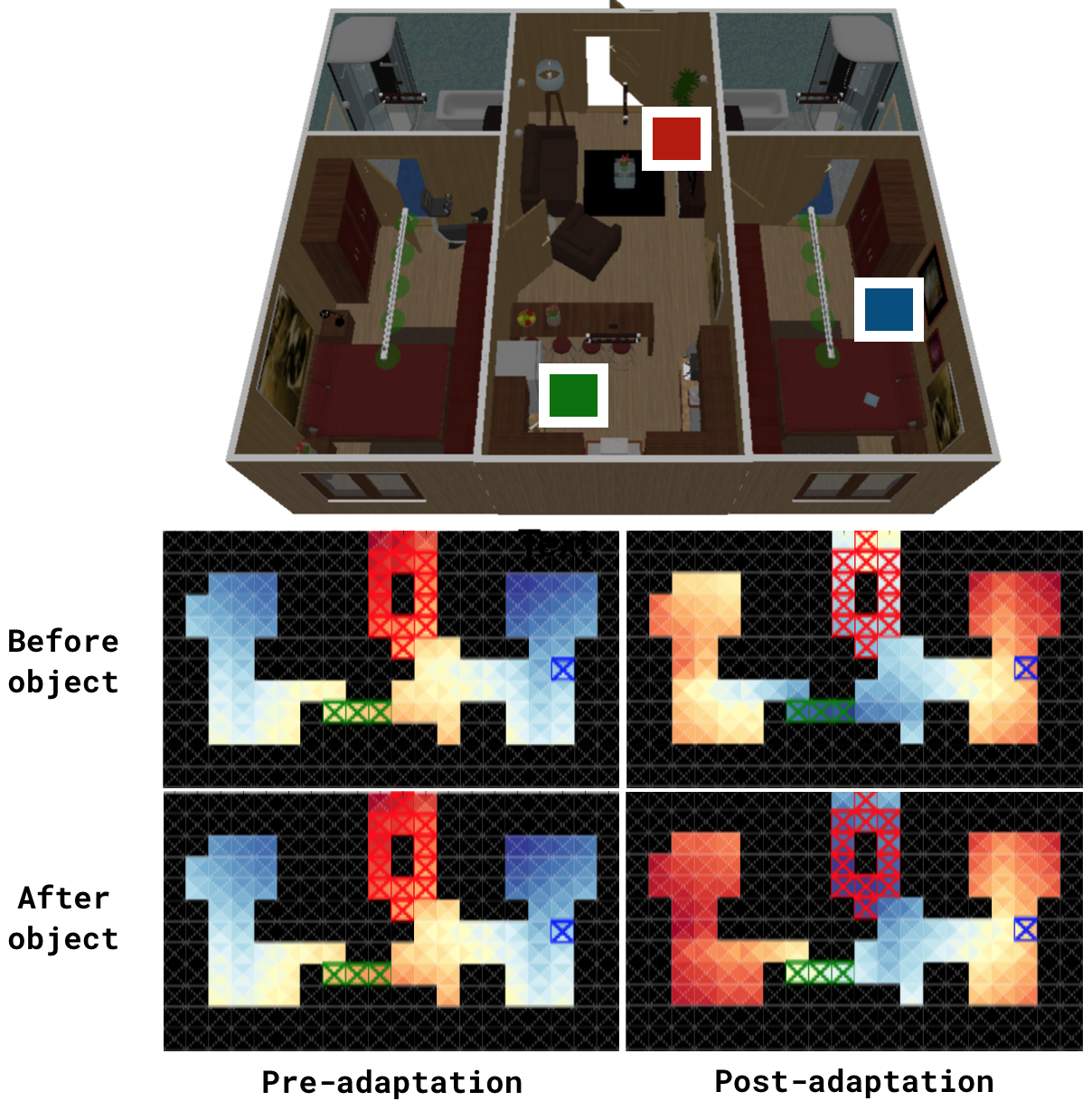} 
    \vspace{-.325cm}
    \captionsetup{font=footnotesize}
    \caption{\small An example adaptation in an \textsc{UNSEEN-HOUSE} (best viewed in color). The agent starts in one room (blue square) and is required to pick up the vase (green square) and take it to the living room (red square). The value function (blue is high, red is low) under the learned reward (bottom) exhibits no ``PICK'' task structure pre-adaptation (bottom left-column). Post-adaptation (bottom right-column), the reward function successfully leads the agent to the vase (bottom figure, top-right plot) and after the pick action (bottom figure, bottom-right plot) is performed, navigates the agent to the goal location.} \label{fig:reward_viz}
    \vspace{-.65cm}
\end{figure}

%% file: wrapfigs/plots.tex
%\newcommand{\plotsize}{0.44\columnwidth}
%\begin{wrapfigure}[37]{R}{\plotsize}
\begin{figure}[h]
    \vspace{-.05in}
    \centering
    \includegraphics[width=1.00\linewidth]{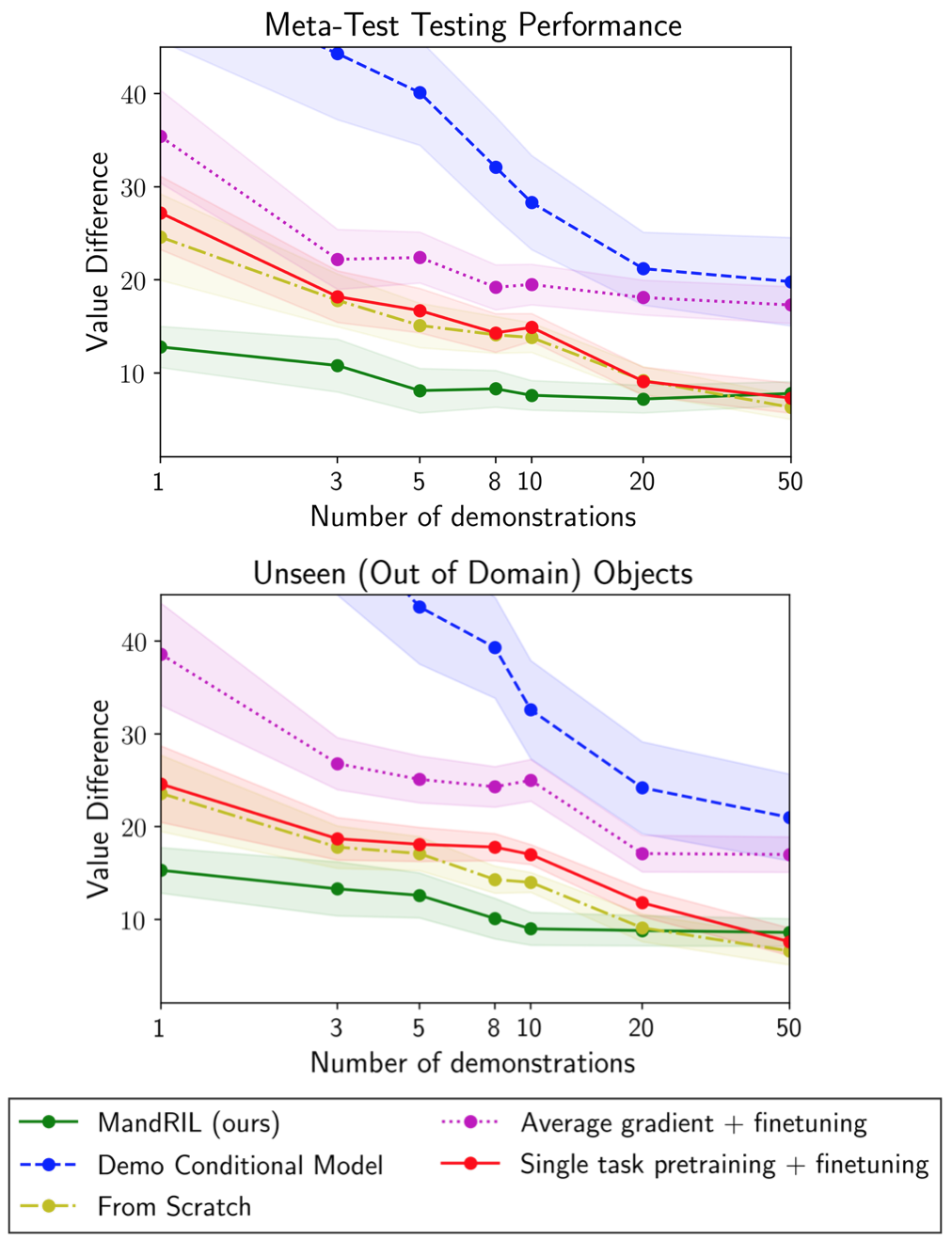}
    \vspace{-.3in}
    \captionsetup{font=footnotesize}
    \caption{\small Meta-test performance on the SpriteWorld domain (lower is better): held-out tasks performance (top) and  held-out tasks with novel sprites (bottom). The recurrent meta-learner has a value difference $>60$ in both test settings. In both test settings, MandRIL achieves comparable performance to the training environment, while the other methods overfit until they receive at least 10 demonstrations (see Appendix \ref{appendix:extra_plot} for training environment performance). We find that pre-training on the full set of tasks leads to negative transfer, while pre-training on a single task is comparable to random initialization. ManDRIL outperforms both alternative initialization approaches, which shows that optimizing for initial weights for fine-tuning robustly improves performance. Shaded regions show 95\% confidence intervals.}
    \label{fig:results}
    \vspace{-.5cm}
\end{figure}

%% file: suncg_table.tex
\begin{table}[t]
\vspace{-0.3cm}
\caption{Success rate (\%) on heldout tasks with 5 demonstrations. ManDRIL achieves consistently better performance on all task/environment types. Results are averaged over 3 random seeds.}
\label{tab:suncg_results}
\vspace{.10cm}
\centering
\resizebox{1.025\columnwidth}{!}{
\begin{sc}
\begin{tabular}{l|c|c|c|c|c|c}
\toprule
\multirow{2}{*}{Method} & \multicolumn{3}{c|}{Test} & \multicolumn{3}{c}{Unseen Houses}  \\
\cline{2-7}
&  Pick & NAV & Total & Pick & Nav & Total \\
\midrule
Behavioral Cloning & ~0.4 & ~8.2  & ~4.3 & ~3.7 & 12.0 & ~9.4 \\
MaxEnt IRL (avg gradient) & 37.3 & 83.7 & 60.8 & 38.3 & 89.7 & 73.3 \\
MaxEnt IRL (from scratch)& 42.4  & 87.9 & 65.4 & 48.1 & 89.9 & 76.5 \\
MandRIL(ours) & \textbf{52.3} &  \textbf{90.7} & \textbf{77.3} & \textbf{56.3} & \textbf{91.0} & \textbf{82.6}\\
%&  Nav & Pick & Total & Nav & Pick & Total \\
%\midrule
%Behavioral Cloning & ~8.2 & ~0.4 & ~4.3 & 12.0 & 3.7 & 9.41 \\
%MaxEnt IRL (pretrained) & 83.7 & 37.3 & 60.8 & 89.7 & 38.3 & 73.3 \\
%MaxEnt IRL (rand init) & 87.9 & 42.4 & 65.4 & 89.9 & 48.1 & 76.5 \\
%MandRIL(ours) & \textbf{90.7} & \textbf{52.3} & \textbf{77.3} & \textbf{91.0} & \textbf{56.3} & \textbf{82.6}\\
\bottomrule
MandRIL (Pre-adaptation) & ~6.0 &  35.3 & 20.7 & ~4.3 & 34.6 & 25.3 \\
\bottomrule
\end{tabular}
\end{sc}
}
\vspace{-0.6cm}
\end{table}

%% file: discussion.tex
\vspace{-0.35cm}
\section{Conclusion}
\vspace{-0.1cm}

In this work, we present an approach that enables few-shot learning of reward functions. We achieve this through a novel formulation of IRL that learns to encode common structure across tasks. Using our meta-IRL approach, we show that we can leverage data from previous tasks to effectively learn reward functions from raw pixel observations for new tasks, from only a handful of demonstrations. Our work paves the way for future work that considers unknown dynamics, or work that employs more fully probabilistic approaches to reward and goal inference.

%% file: appendixa.tex
\mbox{}
\clearpage
\newpage
\appendix
\part*{Appendix}
\label{appendix}
\vspace{-0.5cm}
\section{SpriteWorld Experimental Details}
\label{appendix:spriteworld}
\subsection{Algoritmic Details}
\label{appendix:algo_details}
The input to our reward function for all experiments in this domain is a $80\times80$ RGB image, with an output space of $400$ in the underlying MDP state space. We parameterize the reward function for all methods starting from the same base learner whose architecture we summarize in Table~\ref{table:hyperparams}.

Our LSTM~\citep{hochreiter1997long} implementation is based on the variant used in \citet{zaremba2014recurrent}. The input to the LSTM at each time step is the location of the agent, embedded as the $(x,y)$-coordinates. This is used to predict an spatial map fed as input to the base CNN. We also experimented with conditioning the initial hidden state on image features from a separate CNN, but found that this did not improve performance.

In our demo conditional model, we preserve the spatial information of the demonstrations by feeding in the state visitation map as a image-grid, upsampled with bi-linear interpolation, as an additional channel to the image. In our setup, both the demo-conditional models share the same convolutional architecture, but differ only in how they encode condition on the demonstrations. 

For all our methods, we optimized our model with Adam~\citep{kingma2014adam}. We tuned over the learning rate $\alpha$, the inner learning rate $\beta$ and $\ell_2$ weight decay on the initial parameters. We initialize our models with the Glorot initialization~\cite{glorot2010understanding}. In our LSTM learner, we tuned over embedding sizes and dimensionality. A negative result we found was that bias transformation~\citep{finn2017one} did not help in our experimental setting.
\vspace{-0.9cm}
\begin{table}[H]
\footnotesize
\caption{Hyperparameter summary on Spriteworld environment. Curly brackets indicate the parameter was chosen from that set.}
\label{table:hyperparams}
\begin{center}
\begin{tabular}{@{}ll@{}}
\toprule
Hyperparameters & Value\\
\midrule
Architecture &Conv($256-8\times8-2$)\\
&Conv($128-4\times4-2$)\\
&Conv($64-3\times3-1$)\\
&Conv($64-3\times3-1$)\\
&Conv($1-1\times1-1$)\\
\midrule
Learning rate $\alpha$ & \{0.0001, 0.00001\}\\
Inner learning rate $\beta$ & \{0.001, 0.0005\}\\
Weight decay $\ell_2$ & \{0, 0.0001\}\\
Inner gradient steps & \{1, 3\} \\
Max meta-test gradient steps & \{20\} \\
\midrule
LSTM hidden dimension &\{128, 256\}\\
LSTM embedding sizes & \{64, 128\}\\
\midrule
Batch size & 16 \\
Total meta-training environments & 1000 \\
Total meta-val/test environments & 32\\
Maximum horizon (T) & 15\\
\bottomrule
\end{tabular}
\end{center}
\end{table}
\vspace{-1cm}

\subsection{Environment Details}
\label{appendix:env_details}

The underlying MDP structure of SpriteWorld is a grid, where the states are each of the grid cells, and the actions enable the agent to move to any one of its 8-connected neighbors. The task visuals are inspired by Starcraft (e.g.~\cite{synnaeve2016torchcraft}), although we do not use the game engine. The sprites in our environment are extracted directly from the StarCraft files. We used in total 100 random units for meta-training. Evaluation on new objects was performed with 5 randomly selected sprites. For computational efficiency, we create a meta-training set of 1000 tasks and cache the optimal policy and state visitations under the true cost. Our evaluation is over 32 tasks. Our set of sprites was divided into two categories: buildings and characters. Each characters had multiple poses (taken from different frames of animation, such as walking/running/flying), whereas buildings only had a single pose. During meta-training the units were randomly placed, but to avoid the possibility that the agent would not need to actively avoid obstacles, the units were placed away from the boundary of the image in both the meta-validation and meta-test set.

The terrain in each environment was randomly generated using a set of tiles, each belonging to a specific category (e.g. grass, dirt, water). For each tile, we also specified a set of possible tiles for each of the 4-neighbors. Using these constraints on the neighbors, we generated random environment terrains using a graph traversal algorithm, where successor tiles were sampled randomly from this set of possible tiles. This process resulted in randomly generated, seamless environments. The expert demonstrations were generated using a cost (negative reward) of 8 for the obstacles, 2 for any grass tile, and 1 for any dirt tile. The names of the units used in our experiments are as follows (names are from the original game files):

The list of buildings used is: academy, assim, barrack, beacon, cerebrat, chemlab, chrysal, cocoon, comsat, control, depot, drydock, egg, extract, factory, fcolony, forge, gateway, genelab, geyser, hatchery, hive, infest, lair, larva, mutapit, nest, nexus, nukesilo, nydustpit, overlord, physics, probe, pylon, prism, pillbox, queen, rcluster, refinery, research, robotic, sbattery, scolony, spire, starbase, stargate, starport, temple, warm, weaponpl, wessel. 

The list of characters used is: acritter, arbiter, archives, archon, avenger, battlecr, brood, bugguy, carrier, civilian, defiler, dragoon, drone, dropship, firebat, gencore, ghost, guardian, hydra, intercep, jcritter, lurker, marine, missile, mutacham, mutalid, sapper, scout, scv, shuttle, snakey, spider, stank, tank, templar, trilob, ucereb, uikerr, ultra, vulture, witness, zealot, zergling.

\section{SUNCG Experimental Details}
\label{appendix:suncg}
\subsection{Algorithmic Details}
\label{appendix:suncg_algo_details}
\vspace{-0.2cm}
\begin{table}[H]
\footnotesize
\caption{Hyperparamters on the SUNCG environment. Curly brackets indicate that the the parameter was chosen from that set.}
\label{table:suncg_hyperparams}
\begin{center}
\begin{tabular}{@{}ll@{}}
\toprule
Hyperparameters & Value\\
\midrule
Architecture & Conv($16-5\times5-1$)\\
&Conv($32-3\times3-1$)\\
&MLP($32$)\\
&MLP($1$)\\
\midrule
Max number of training steps & 15000000 \\
Number of seed & 3 \\
\midrule
Learning rate $\alpha$ & \{0.1, 0.01, 0.001, 0.0001\} \\
Inner learning rate $\beta$ & \{0.15, 0.1, 0.01, 0.0001\}\\ 
Inner gradient steps & \{3, 5\} \\
Max meta-test gradient steps & \{10\} \\
Momentum & \{0.9, 0.95, 0.99\} \\
\bottomrule
\end{tabular}
\end{center}
\end{table}
\vspace{-0.25cm}

Our per task MaxEnt IRL baseline is learned by using the same base architecture. To provide a fair comparison, we do not use an inner learning rule in the inner loop of ManDRIL such as Adam~\cite{kingma2014adam} and use regular SGD. For our baseline however, we include a momentum term over which we tune. We tune over the number of training steps, learning rate and momentum parameters. We use SGD with momentum. For ManDRIL, we tune over the inner learning rate $\beta$ and learning rate $\alpha$ and number of gradient steps. At meta-test time, we experimented with taking up to $10$ gradient steps. For pretraining IRL, we first train for 150,000 steps, freeze the weights, and fine tune them for every separate task. For training from scratch, we use the Glorot uniform initialization in the the convolutional layers~\cite{glorot2010understanding}. 

\subsection{Environment Details}
\label{appendix:suncg_env_details}
\vspace{-0.2cm}
\begin{table}[H]
\footnotesize
\caption{Summary of SUNCG environment setup.}
\label{table:suncg_env_params}
\begin{center}
\begin{tabular}{@{}ll@{}}
\toprule
Hyperparameters & Value\\
\midrule
Discount ($\gamma$) & 0.99\\
Maximum horizon (T) & 40\\
Initial random steps & 30 \\
Number of demonstrations & 5 \\
\midrule
Training environments & 1004 \\
Test environments & 236\\
Test-house environments & 173\\
(PICK/NAV) split: & 716/697\\
\bottomrule
\end{tabular}
\end{center}
\end{table}
\vspace{-0.25cm}

The MDP in each environment is discretized into a grid where the state is defined by the grid coordinates plus the agent's orientation (N,S,E,W). The agent receives an observation which is a first-person panoramic view. The panoramic view consists of four $32\times24$ semantic image observations containing $61$ channels. 

The only departure for the task setup of \citet{fu2019language} that we make is to randomize the agent's start location by executing a random walk at the beginning of each episode. In \citet{fu2019language}, the agent's start location was previously deterministic which allows a trivial solution of memorizing the provided demonstrations.

\section{SpriteWorld Meta-Test Training Performance}
\label{appendix:extra_plot}
\begin{figure}[H]
    \vspace{-0.25cm}
    \centering
    \includegraphics[width=0.85\columnwidth]{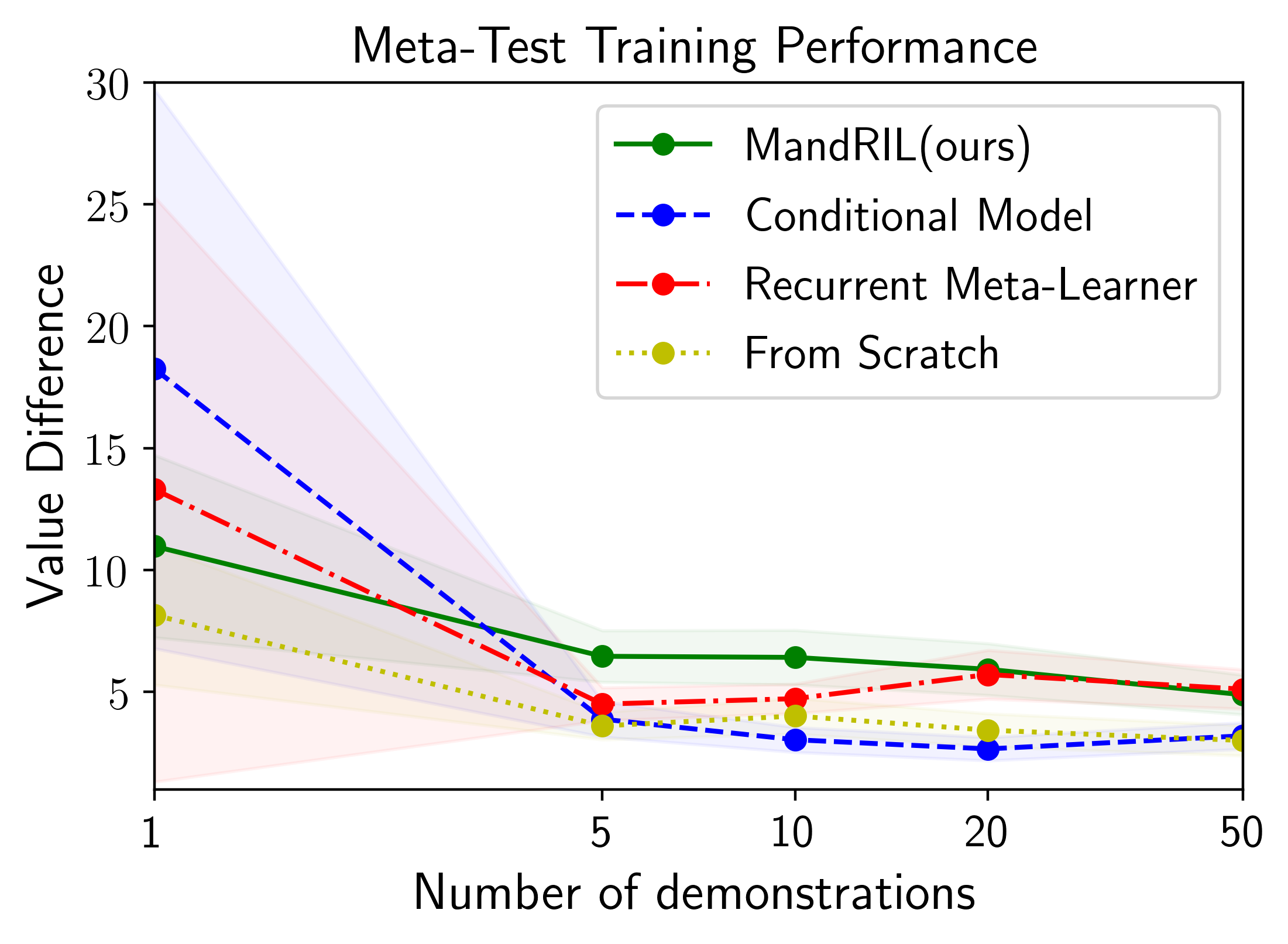}
    \label{fig:meta-train}
    \captionsetup{font=footnotesize}
    \caption{\small Meta-test ``training'' performance with varying numbers of demonstrations (lower is better). This is the performance on the environment for which demonstrations are provided for adaptation. As the number of demonstrations increase, all methods are able to perform well in terms of training performance as they can simply overfit to the training environment without acquiring the right visual cues that allow them to generalize. However, we find comes at the cost comes of considerable overfitting as we discuss in Section.~\ref{sec:experiments}.}
\end{figure}
\mbox{}

\section{SUNCG DAgger Performance}
\label{appendix:dagger}
\input{suncg_dagger.tex}

Here we show the performance of DAgger~\citep{ross2011reduction}, in the setting where the number of samples that is equal to the number of demonstrations. Overall, while DAgger slightly improves performance over behavioral cloning, the performance still lags significantly behind ManDRIL and other IRL methods.

\clearpage
\newpage
\section{Detailed Meta-Objective Derivation}
\label{appendix:derivation}
We define the quality of reward function $r_{\vtheta}$ parameterized by $\vtheta \in \mathbb{R}^k$ on task $\mathcal{T}$ with the MaxEnt IRL loss, $\mathcal{L}_{\text{IRL}}^{\mathcal{T}}(\vtheta)$, described in Section~\ref{sec:methods}. The corresponding gradient is
\begin{equation} \label{eq:grad_max_ent}
    \nabla_{\vtheta} \mathcal{L}_{\text{IRL}}(\vtheta) = \frac{\partial\,r_{\vtheta}}{\partial\,\vtheta} ( \mathbb{E}_\tau [\vmu_\tau] - \vmu_{\mathcal{D}_{\mathcal{T}}} ),
\end{equation}
where $\partial\,r_{\vtheta}/\partial\,\vtheta$ is the $k \times |\mathcal{S}| |\mathcal{A}|$-dimensional Jacobian matrix of the reward function $r_{\vtheta}$ with respect to the parameters $\vtheta$. Here, $\vmu_\tau \in \mathbb{R}^{\vert \mathcal{S} \vert  \vert \mathcal{A} \vert}$ is the vector of \emph{state-action visitations} under the trajectory $\tau$ (i.e. the vector whose elements are 1 if the corresponding state-action pair has been visited by the trajectory $\tau$, and 0 otherwise), and $\vmu_{\mathcal{D}_{\mathcal{T}}} = \frac{1}{\vert \mathcal{D}_{\mathcal{T}} \vert} \sum_{\tau \in \mathcal{D}_{\mathcal{T}}} \vmu_\tau$ is the mean state visitations over all demonstrated trajectories in $\mathcal{D}_{\mathcal{T}}$. Let $\vphi_{\mathcal{T}} \in \mathbb{R}^k$ be the updated parameters after a single gradient step. Then
\begin{equation} \label{eq:phi_T}
    \vphi_{\mathcal{T}} = \vtheta - \alpha \nabla_{\vtheta} \tasklosstrain(\vtheta).
\end{equation}

Let $\tasklosstest$ be the MaxEnt IRL loss, where the expectation over trajectories is computed with respect to a test set that is \emph{disjoint} from the set of demonstrations used to compute $\tasklosstest(\vtheta)$ in Eq.~\ref{eq:phi_T}. We seek to minimize
\begin{equation} \label{eq:meta_irl_obj}
    \sum_{\mathcal{T} \in \mathcal{T}^{\text{test}}} \tasklosstest (\vphi_{\mathcal{T}})
\end{equation}
over the parameters $\vtheta$. To do so, we first compute the gradient of Eq.~\ref{eq:meta_irl_obj}, which we derive here. Applying the chain rule
\begin{align} \label{eq:tmp}
    &\nabla_{\vtheta} \tasklosstest  = \frac{\partial\,\vphi_{\mathcal{T}}}{\partial\,\vtheta} \frac{\partial\,r_{\vphi_{\mathcal{T}}}}{\partial\,\vphi_{\mathcal{T}}} \frac{\partial\,\tasklosstest}{\partial\,r_{\vphi_{\mathcal{T}}}} \nonumber \\ 
    & = \frac{\partial\,}{\partial\,\vtheta}\left( \vtheta - \alpha \nabla_{\vtheta}  \tasklosstrain(\vtheta) \right)  \frac{\partial\,r_{\vphi_{\mathcal{T}}}}{\partial\,\vphi_{\mathcal{T}}} \frac{\partial\,\tasklosstest}{\partial\,r_{\vphi_{\mathcal{T}}}} \nonumber \\ 
    & = \left( \mathbf{I} - \alpha \frac{\partial\,}{\partial\,\vtheta}\left( \frac{\partial\,r_{\vtheta}}{\partial\,\vtheta} ( \mathbb{E}_\tau [\vmu_\tau] - \vmu_{\mathcal{D}_{\mathcal{T}}} ) \right)\right) \frac{\partial\,r_{\vphi_{\mathcal{T}}}}{\partial\,\vphi_{\mathcal{T}}} \frac{\partial\,\tasklosstest}{\partial\,r_{\vphi_{\mathcal{T}}}}  
\end{align}
where in the last equation we substitute in the gradient of the MaxEnt IRL loss in Eq.~\ref{eq:grad_max_ent} for $\nabla_{\vtheta}  \tasklosstrain(\vtheta)$. In Eq.~\ref{eq:tmp}, we use the following notation:
\begin{itemize}
    \item $\partial\,\vphi_{\mathcal{T}}/\partial\,\vtheta$ denotes the $k\times k$-dimensional vector of partial derivatives $\partial\,\vphi_{\mathcal{T}, i}/\partial\,\vtheta_j$,
    \item $\partial\,r_{\vphi_{\mathcal{T}}}/\partial\,\vphi_{\mathcal{T}}$ denotes the $k\times |\mathcal{S}| |\mathcal{A}|$-dimensional matrix of partial derivatives $\partial\,r_{\vphi_{\mathcal{T}, i}}/\partial\,\vphi_{\mathcal{T}, j}$,
    \item and, $\partial\,\tasklosstest/\partial\,r_{\vphi_{\mathcal{T}}}$ denotes the $k$-dimensional gradient vector of $\tasklosstest$ with respect to $r_{\vphi_{\mathcal{T}}}$.
\end{itemize}

We will now focus on the term inside of the parentheses in Eq.~\ref{eq:tmp}, which is a $k \times k$-dimensional matrix of partial derivatives. 

%\begin{align*}
%    \frac{\partial\,}{\partial\,\vtheta}\left( \frac{\partial\,r_{\vtheta}}{\partial\,\vtheta} ( \mathbb{E}_\tau [\vmu_\tau] - \vmu_{\mathcal{D}_{\mathcal{T}}} ) \right) & = \sum_{i=1}^{|\mathcal{S}| |\mathcal{A}|} \frac{\partial^2\,r_{\vtheta}}{\partial\,\vtheta^2} (\mathbb{E}_\tau [\vmu_\tau] - \vmu_{\mathcal{D}_{\mathcal{T}}})_i + \left(\frac{\partial\,}{\partial\,\vtheta}\mathbb{E}_\tau [\vmu_\tau]\right) \left(\frac{\partial\,r_{\vtheta}}{\partial\,\vtheta}\right)^\top \\
%    & = \sum_{i=1}^{|\mathcal{S}| |\mathcal{A}|} \frac{\partial^2\,r_{\vtheta}}{\partial\,\vtheta^2} (\mathbb{E}_\tau [\vmu_\tau] - \vmu_{\mathcal{D}_{\mathcal{T}}})_i + \left( \frac{\partial\,r_{\vtheta}}{\partial\,\vtheta}  \right)\left(\frac{\partial\,}{\partial\,r_{\vtheta}}\mathbb{E}_\tau [\vmu_\tau]\right) \left(\frac{\partial\,r_{\vtheta}}{\partial\,\vtheta}\right)^\top
%\end{align*}

% ER: Edited 11/12/2018 to reflect reviewer's corrections
\begin{align*}
    &\frac{\partial\,}{\partial\,\vtheta}\left( \frac{\partial\,r_{\vtheta}}{\partial\,\vtheta} ( \mathbb{E}_\tau [\vmu_\tau] - \vmu_{\mathcal{D}_{\mathcal{T}}} ) \right) \\
    &= \sum_{i=1}^{|\mathcal{S}| |\mathcal{A}|} \left[ \frac{\partial^2\,r_{\vtheta}}{\partial\,\vtheta^2} (\mathbb{E}_\tau [\vmu_\tau] - \vmu_{\mathcal{D}_{\mathcal{T}}})_i + \frac{\partial\,}{\partial\,\vtheta}(\mathbb{E}_\tau [\vmu_\tau])_i \left(\frac{\partial\,r_{\vtheta, i}}{\partial\,\vtheta}\right)^\top \right] \\
    \begin{split}
    &= \sum_{i=1}^{|\mathcal{S}| |\mathcal{A}|} \left[ \frac{\partial^2\,r_{\vtheta}}{\partial\,\vtheta^2} (\mathbb{E}_\tau [\vmu_\tau] - \vmu_{\mathcal{D}_{\mathcal{T}}})_i + \right.\\
    &\left( \frac{\partial\,r_{\vtheta, i}}{\partial\,\vtheta}  \right)\left(\frac{\partial\,}{\partial\,r_{\vtheta, i}}(\mathbb{E}_\tau [\vmu_\tau])_i\right) \left(\frac{\partial\,r_{\vtheta, i}}{\partial\,\vtheta}\right)^\top \left. \right]
    \end{split}
\end{align*}

where between the first and second lines, we apply the chain rule to expand the second term. In this expression, we make use of the following notation:
\begin{itemize}
    \item $\partial^2\,r_{\vtheta}/\partial\,\vtheta^2$ denotes the $k\times |\mathcal{S}| |\mathcal{A}|$-dimensional matrix of second-order partial derivatives of the form $\partial^2\,r_{\vtheta, i}/\partial\,\vtheta_j^2$,
    \item $(\mathbb{E}_\tau [\vmu_\tau] - \vmu_{\mathcal{D}_{\mathcal{T}}})_i$ denotes the $i$th element of the $|\mathcal{S}| |\mathcal{A}|$-dimensional vector $(\mathbb{E}_\tau [\vmu_\tau] - \vmu_{\mathcal{D}_{\mathcal{T}}})_i$,
    \item $\partial\,r_{\vtheta, i}/\partial\,\vtheta$ denotes the $k$-dimensional matrix of partial derivatives of the form $\partial\,r_{\vtheta, i} / \partial\,\vtheta_j$ for $j = 1, 2, \dots, k$,
    \item and, $\frac{\partial\,}{\partial\,r_{\vtheta, i}}(\mathbb{E}_\tau [\vmu_\tau])_i$ is the partial derivative of the $i$th element of the $|\mathcal{S}| |\mathcal{A}|$-dimensional vector $\mathbb{E}_\tau [\vmu_\tau]$ with respect to the $i$th element of the $|\mathcal{S}| |\mathcal{A}|$-dimensional vector $r_{\vtheta}$ of reward (i.e. the reward function). 
\end{itemize}

When substituted back into Eq.~\ref{eq:tmp}, the resulting gradient is equivalent to that in Eq.~\ref{eq:grad_meta_obj} in Section~\ref{sec:methods}. In particular, defining the $|\mathcal{S}| |\mathcal{A}|$-dimensional diagonal matrix $D$ as 
\[ D := \text{diag}\left(\left\{\frac{\partial\,}{\partial\,r_{\vtheta, i}}(\mathbb{E}_\tau [\mu_\tau])_i\right\}_{i=1}^{|\mathcal{S}| |\mathcal{A}|}\right) \]
then the final term can be simplified to
\begin{align*}
&\sum_{i=1}^{|\mathcal{S}| |\mathcal{A}|} \left( \frac{\partial\,r_{\vtheta, i}}{\partial\,\vtheta}  \right)\left(\frac{\partial\,}{\partial\,r_{\vtheta, i}}(\mathbb{E}_\tau [\vmu_\tau])_i\right) \left(\frac{\partial\,r_{\vtheta, i}}{\partial\,\vtheta}\right)^\top \\
&= \left(\frac{\partial\,r_{\vtheta}}{\partial\,\vtheta}\right)D\left(\frac{\partial\,r_{\vtheta}}{\partial\vtheta}\right)^\top.
\end{align*}
In order to compute this gradient, however, we must take the gradient of the expectation $\mathbb{E}_\tau [\vmu_\tau]$ with respect to the reward function $r_{\vtheta}$. This can be done by expanding the expectation as follows
\begin{align*}
    &\frac{\partial\,}{\partial\,r_{\vtheta}}\mathbb{E}_\tau [\vmu_\tau] = \frac{\partial\,}{\partial\,r_{\vtheta}} \sum_\tau \left(\frac{\exp(\vmu_\tau^\top r_{\vtheta})}{\sum_{\tau'} \exp(\vmu_{\tau'}^\top r_{\vtheta})}\right) \vmu_\tau \\
    & = \sum_\tau \left(  \left(\frac{\exp(\vmu_\tau^\top r_{\vtheta})}{\sum_{\tau'} \exp(\vmu_{\tau'}^\top r_{\vtheta})}\right)  (\vmu_\tau\vmu_\tau^\top) - \frac{\exp(\vmu_\tau^\top r_{\vtheta})}{(\sum_{\tau'} \exp(\vmu_{\tau'}^\top r_{\vtheta}))^2} \sum_{\tau'} (\vmu_{\tau'}\vmu_\tau^\top)\exp(\vmu_{\tau'}^\top r_{\vtheta}) \right) \\
    & = \sum_\tau P(\tau \mid r_{\vtheta})(\vmu_\tau \vmu_\tau^\top) - \sum_\tau P(\tau | r_{\vtheta}) \sum_{\tau'} P(\tau' \mid r_{\vtheta}) ( \vmu_{\tau'} \vmu_\tau^\top) \\
    & = \mathbb{E}_\tau\left[(\vmu_\tau \vmu_\tau^\top) - \sum_{\tau'} P(\tau' \mid r_{\vtheta}) ( \vmu_{\tau'} \vmu_\tau^\top)\right] \\
    & = \mathbb{E}_\tau[\vmu_\tau \vmu_\tau^\top] - \mathbb{E}_{\tau', \tau}[\vmu_{\tau'}\vmu_\tau^\top] \\
    & = \mathbb{E}_\tau[\vmu_\tau \vmu_\tau^\top] - \mathbb{E}_\tau [\vmu_\tau](\mathbb{E}_\tau [\vmu_\tau])^\top \\
    & = \text{Cov}[\vmu_\tau].
\end{align*}

%% file: suncg_dagger.tex
\begin{table}[!h]
\vspace{-0.5cm}
\caption{DAgger success rate (\%) on heldout tasks with 5 demonstrations. ManDRIL values are repeat for viewing convenience. Results are averaged over 3 random seeds.}
\label{tab:suncg_results}
\vspace{.10cm}
\centering
\resizebox{1.025\columnwidth}{!}{
\begin{sc}
\begin{tabular}{l|c|c|c|c|c|c}
\toprule
\multirow{2}{*}{Method} & \multicolumn{3}{c|}{Test} & \multicolumn{3}{c}{Unseen Houses}  \\
\cline{2-7}
&  Pick & NAV & Total & Pick & Nav & Total \\
\midrule
DAgger & ~1.0 & 12.8  & ~7.5 & ~7.4 & 15.5 & 11.8 \\
MandRIL(ours) & \textbf{52.3} &  \textbf{90.7} & \textbf{77.3} & \textbf{56.3} & \textbf{91.0} & \textbf{82.6}\\
\bottomrule
\end{tabular}
\end{sc}
}
\end{table}